\documentclass[12pt]{article}
\usepackage{fancyhdr}
\usepackage{xcolor}
\usepackage{array}
\usepackage{hyperref}
\usepackage{ulem}
\usepackage[english]{babel}
\usepackage[nottoc]{tocbibind}
\usepackage{csquotes}
\usepackage[toc,page]{appendix}
\usepackage{booktabs}
\newtheorem{theorem}{Theorem}

\hypersetup{
   colorlinks=true,
   linkcolor=cyan,
   filecolor=green,
   urlcolor=blue
}
\usepackage{graphicx,amssymb,amsmath,multicol,nicefrac} %% figure package
\usepackage{comment}

\providecommand{\keywords}[1]{\textbf{Keywords---} #1}

\usepackage[sorting=none]{biblatex}

\usepackage{authblk}
\usepackage{blindtext}

\begin{document}

\title{AI Alignment and Social Choice: Fundamental Limitations and Policy Implications\footnote{I thank Bhasi Nair for discussions and comments on an earlier draft of this paper. This research was made possible through generous support from the Kevin Xu Initiative on Science, Technology, and Global Development at the University of Chicago and by the Patrick J. McGovern Foundation at Equitech Futures.}}
\author{Abhilash Mishra %\href{https://orcid.org/0000-0000-0000-0000}{\includegraphics[scale=0.04]{orcidicon.eps}}
\thanks{Email: abhilash@equitechfutures.com, abhilashmishra@uchicago.edu}}
\affil{Equitech Futures \& The University of Chicago}

\maketitle
\vspace{6pt}

\vspace{6pt}

\begin{abstract}
\noindent Aligning AI agents to human intentions and values is a key bottleneck in building safe and deployable AI applications. But whose values should AI agents be aligned with? Reinforcement learning with human feedback (RLHF) has emerged as the key framework for AI alignment. RLHF uses feedback from human reinforcers to fine-tune outputs; all widely deployed large language models (LLMs) use RLHF to align their outputs to human values. It is critical to understand the limitations of RLHF and consider policy challenges arising from these limitations. In this paper, we investigate a specific challenge in building RLHF systems that respect democratic norms. Building on impossibility results in social choice theory, we show that, under fairly broad assumptions, there is {\it no unique voting protocol} to universally align AI systems using RLHF through democratic processes. Further, we show that aligning AI agents with the values of all individuals will always violate certain private ethical preferences of an individual user i.e. universal AI alignment using RLHF is impossible. We discuss policy implications for the governance of AI systems built using RLHF: first the need for mandating transparent voting rules to hold model builders accountable. Second, the need for model builders to focus on developing AI agents that are narrowly aligned to specific user groups. 
\end{abstract}

\newpage

\section{Introduction}

The challenge of ensuring that artificial intelligence (AI) systems are aligned with human intentions, ethics, and values has been recognized since the early days of the development of AI. In his seminal article {\it Some Moral and Technical Consequences of Automation}, Norbert Wiener~\cite{wiener1960some} highlighted the need to ensure that the ``purpose put into the machine is the purpose which we really desire." More recently, Russell~\cite{russell2021human} has highlighted the need to explicitly align AI with objectives aligned with human intentions and values. 

\medskip

\noindent But how can we embed human intentions and values in machines? And whose values should be embedded? With the rise of autonomous systems, these questions have garnered significant attention in recent years. This has led to new disciplines like {\it Machine Behavior}~\cite{rahwan2019machine} and {\it Machine Ethics}~\cite{anderson2011machine} which have explored how machines can be ``taught" to learn human norms. As expected, there is no unique way to teach norms to machines. Building on Rawlsian ideas of fairness~\cite{rawls2001justice}, Dwork et al.~\cite{dwork2012fairness} suggested that machine norms need to be ``an approximation as agreed upon by society.” Awad et al.~\cite{awad2018moral} operationalized this idea for the case of autonomous vehicles by crowd-sourcing moral decisions from millions of online participants through the Moral Machine Experiment. Noothigattu et al.~\cite{noothigattu2018voting} used data from the Moral Machine Experiment to build a model of aggregated moral preferences using tools from computational social choice~\cite{brandt2012computational, brandt2016handbook}.

\medskip

\noindent More recently, in the context of Large Language Models (LLMs), the project of AI alignment has been focused on mitigating harms i.e. ensuring that LLMs follow user instructions, do not generate biased or toxic text, or factually inaccurate information~\cite{askell2021general}. Reinforcement learning using human feedback (RLHF) has been the key technical innovation that has led to remarkable progress in developing aligned AI agents~\cite{bai2022training, christiano2017deep, ziegler2019fine}. For instance, the most widely deployed LLMs like GPT4~\cite{openai2023gpt4}, Claude~\cite{anthropic2023introducing}, Bard~\cite{thoppilan2022lamda}, and Llama~\cite{touvron2023llama} all use RLHF to fine-tune their models. While these approaches focus on criteria such as ``helpfulness, honesty, and harmlessness"~\cite{askell2021general}, there remain open questions about how such alignment can be realized practically and at scale. More broadly, whose norms or values should they be aligned with? And how can we align AI systems by respecting democratic norms?

\medskip

\noindent The limitations of RLHF in aligning AI systems at scale remain poorly understood. Casper et. al.~\cite{casper2023open} provide an exhaustive review of fundamental limitations and open challenges in RLHF for AI alignment. In particular, they highlight the limitation of selecting ``representative humans" to act as reinforcers during the training process. For example, while training GPT4, OpenAI restricts human reinforcers who agree with ``expert" researcher preferences~\cite{ouyang2022training}. Given systemic issues in researcher training and hiring, this can lead to biased outcomes. Additionally, the demographics of reinforcers are not representative of end users.\footnote{OpenAI reinforcers consist of $\sim$50\% Filipino and Bangladeshi nationals~\cite{ouyang2022training} a composition that is likely driven by low costs of data labour in developing countries.}

\medskip

\noindent This paper examines the limits of building aligned AI systems through a democratic process, {\it even when reinforcers are representative of the underlying population of users.} More specifically, we ask whether it is possible to design voting rules that allow a group of {\it reinforcers}, representative of a population of diverse {\it users}, to train an AI model using RLHF. Using two widely known results from social choice theory, we show that there exist no unique voting rules that allow a group of reinforcers to build an aligned AI system by respecting democratic norms i.e. by treating all users and reinforcers the same. Further, we show that it is impossible to build a RLHF model democratically that respects the {\it private preferences} of each user in a population simultaneously. As far as we know, this fundamental limitation of RLHF has not been highlighted in the literature.

\medskip

\noindent This paper builds on ideas from social choice theory, machine ethics, and computer science to highlight some fundamental barriers to embedding human intentions to AI systems while following democratic norms. Our key result borrows from two widely known theorems in social choice theory - the impossibility theorems by Arrow and Sen - which are widely known constraints in voting theory. Gabriel~\cite{gabriel2020artificial} briefly touched on the implications of impossibility theorems on AI alignment but as far as we know, these arguments have not been formally investigated in the context of RLHF. 

\medskip

\noindent The rest of the paper is organized as follows. In Section~\ref{sec:RLHF}, we provide formalize reinforcement learning with multiple reinforcers, in Section~\ref{sec:theorems} we outline the Arrow-Sen impossibility theorems in the context of RLHF and we discuss the implications of these theorems in Section~\ref{sec:implications}.

\section{Reinforcement Learning with Multiple Reinforcers}
\label{sec:RLHF}
Early approaches to reinforcement learning explicitly specified a reward function~\cite{mnih2015human} that was used to train an AI model. However, specifying a well-defined reward function to achieve alignment for AI language agents is impossible. Reinforcement Learning with Human Feedback (RLHF)~\cite{christiano2017deep, akrour2014} was suggested as a way to ``communicate" human intentions to language models through a small group of non-expert human reinforcers. The key innovation in RLHF is training AI agents to be aligned with humans {\it without} knowing the explicit reward function. Instead, the reward function is ``discovered" via human feedback.

\medskip

\noindent Consider an AI agent receiving feedback from a human reinforcer over a sequence of steps; at each time $t$ the agent receives a question/prompt $p_t\in\mathcal{O}$ from the human and then sends an answer/response $a_t \in \mathcal{A}$ to the human. The reinforcement function is not known {\it a priori}. Instead, the human reinforcer expresses its {\it preferences} between trajectory segments. A trajectory segment is a sequence of observations and
actions, $\sigma = ((p_0,a_0),(p_1,a_1),...,(p_{k-1},a_{k-1}))\in (\mathcal{O}\times \mathcal{A})^k$. When a human reinforcer prefers  trajectory segment $\sigma_1$ to trajectory segment $\sigma_2$, we denote it as $\sigma_1\succ\sigma_2$. 

\medskip

\noindent Consider a set of trajectory segments $\sigma \in \mathcal{X}$ and a set of human reinforcers $\mathcal{N}$. Each human reinforcer $i\in \mathcal{N}$ ranks the trajectory segments in $\mathcal{X}$ according to an individual preference ranking denoted by $\succeq_i$. The preferences $\succ$ between two trajectory segments $\sigma_1$ and $\sigma_2$ are generated by a reward function $r_i$ for $i\in \mathcal{N}$, such that:
\begin{equation}
    ((p^1_0,a^1_0),(p^1_1,a^1_1),...,(p^1_{k-1},a^1_{k-1})) \succ  ((p^2_0,a^2_0),(p^2_1,a^2_1),...,(p^2_{k-1},a^2_{k-1})) 
\end{equation}
whenever,
\begin{equation}
    r_i(p^1_0,a^1_0)+...+r_i(p^1_{k-1},a^1_{k-1})> r_i(p^2_0,a^2_0)+...+r_i(p^2_{k-1},a^2_{k-1}) 
\end{equation}

\medskip

\noindent If reinforcer $i$ likes trajectory $\sigma_1$ as much as or more than alternative $\sigma_2$, we write $\sigma_1 \succeq \sigma_2$. If they like $\sigma_1$ strictly more than $\sigma_2$ we write $\sigma_1 \succ \sigma_2$, and if they are indifferent we write $\sigma_1 \equiv \sigma_2$. Each reinforcer $i$ is rational i.e. the preferences are complete ($\sigma_1 \succ \sigma_2$ or both) and transitive (if $\sigma_1 \succ \sigma_2$ and $\sigma_2 \succ \sigma_3$ then $\sigma_1 \succ \sigma_3$). 

\medskip

\noindent Thus, the ranking $\succeq_i$ is a weak ordering of the alternatives in $\mathcal{X}$, and the set of all weak orderings of $\mathcal{X}$ is denoted by $\mathcal{R}$. Finally, we denote the profile of all individuals’ preferences by $\rho= (\succeq_1, \succeq_2, ... , \succeq_n)$. A preference aggregation function $f(\rho)$ takes a preference profile $\rho$ as input and produces a collective preference relation, $\succeq$, that compares the alternatives.

\section{Arrow-Sen Impossibility Theorems for RLHF}
\label{sec:theorems}
Consider a set of reinforcers $i\in\mathcal{N}$ who have to vote on their preferences  for a set of trajectories $\sigma_1, \sigma_2$ and $\sigma_3$. There can be multiple voting rules to choose a preference. For example, a {\it plurality rule} would choose a preference that has the largest number of votes, even when a preference does not have a clear majority ($>50\%$) votes. Another alternative is the {\it simple majority rule} that which chooses a preference with the majority of votes in a ranked choice. The two voting rules are shown in Tables~\ref{simple_majority} and~\ref{plurality}. The two simple voting rules above lead to different outcomes: for the plurality rule selects trajectory $\sigma_1$, whereas the majority rule chooses trajectory $\sigma_2$. One can similarly design many other voting rules. 

\medskip

\noindent In the above scenario, which among all possible voting rules is the ``best"? Arrow~\cite{arrow2012social} showed that under very general considerations, no such voting rule exists. While this is a widely known result, its application to RLHF has important implications for building scalable and aligned AI systems. We outline Arrow's arguments below in the context of RLHF. 

\begin{table}
  \centering
  \begin{tabular}{lll}
    \toprule
    30 \%     & 45\%     & 25\% \\
    \midrule
    $\sigma_1$ & $\sigma_2$  & $\sigma_3$     \\
    \bottomrule
  \end{tabular}
\caption{Voting Rules: Simple Majority Rule}
\label{simple_majority}
\end{table}

\begin{table}
  \centering
  \begin{tabular}{lll}
    \toprule
    30 \%     & 45\%     & 25\% \\
    \midrule
    $\sigma_1$ & $\sigma_2$  & $\sigma_3$     \\
    $\sigma_2$ & $\sigma_3$  & $\sigma_1$     \\
    $\sigma_3$ & $\sigma_1$  & $\sigma_2$     \\
    \bottomrule
  \end{tabular}
  \caption{Voting Rules: Plurality Rule}
  \label{plurality}
\end{table}

\medskip

\noindent Arrow~\cite{arrow2012social} posited four axioms that any reasonable aggregation/voting rule should satisfy. The axioms are:
\begin{itemize}
    \item {\bf Pareto or Consensus:} an aggregation function $f$ satisfies Pareto if, whenever every individual $i\in \mathcal{N}$ strictly prefers $\sigma_1$ to $\sigma_2$, the function $f$ ranks $\sigma_1$ strictly higher than $\sigma_2$.
    \item {\bf Independence of irrelevant alternatives (IIA):} the group members’ preferences about some alternative $\sigma_3$ does not affect how the aggregation rule $f$ ranks two different alternatives, $\sigma_1 \neq \sigma_3$ and $\sigma_2\neq \sigma_3$.
    \item {\bf Transitivity:} If $f$ produces an ordering in which $\sigma_1 \succeq \sigma_2$ and $\sigma_2 \succeq \sigma_3$, then it must also be the case that $\sigma_1 \succeq \sigma_3$.
    \item {\bf No Dictator:} a rule $f$ is dictatorial when there exists one voter $i$ such that every time $\sigma_1 \succ_i \sigma_2$ (i.e., the dictator $i$ strictly prefers $\sigma_1$ to $\sigma_2$), the aggregation rule $f$ produces a strict ranking $\sigma_1 \succ \sigma_2$. An aggregation rule $f$ satisfies no dictator if it is not dictatorial.
\end{itemize}
Arrow's impossibility theorem states that no aggregation rule can simultaneously satisfy all four axioms listed above. 

\begin{theorem}[Arrow's Theorem]
With three or more alternatives, any aggregation rule $f$ satisfying Pareto, independence of irrelevant alternatives, and transitivity must be dictatorial.
\end{theorem}

\noindent To understand the implications of Arrow's theorem for RLHF, suppose we want to design an AI agent through a democratic process. Arrow’s theorem implies that {\it any} voting rule that is Pareto efficient, transitive, and independent of irrelevant alternatives must grant all decision-making authority to a single individual/reinforcer.\footnote{The no dictator axiom is a weak constraint. If instead of one reinforcer, two reinforcers have all the power in determining the preference, we would still consider the process undemocratic even if the no dictator axiom is formally satisfied. Truly democratic voting procedures instead require that all reinforcers have the same weight.} 

\medskip

\noindent Sen~\cite{sen1970impossibility} extended the idea of Arrow's theorem to understand the implications for individual rights in a society where decisions are made by the preferences of the majority. Consider, for instance, an individual $i$’s rights within a social choice framework. The individual can choose between two alternatives, $\sigma_1$ and $\sigma_2$, and the two alternatives differ only with respect to features that are private to $i$ i.e. other individuals {\it should not} have a say in $i$'s preference between two alternatives. In his original paper Sen summarized this impossibility result in the following way: ``given other things in the society, if you prefer to have pink walls rather than white, then society should permit you to have this, even if a majority of the community would like to see your walls white." In the case of RLHF, this can be a specific definition of ``harmlessness" that individual $i$ cares about (e.g. expecting the AI system to avoid certain stereotypes based on gender or race). 

\medskip

\noindent Now consider an individual’s ``protected domain" $\mathcal{D}_i=(\sigma_1,\sigma_2)$, which is a collection of pairs of alternatives between which that person’s preference should be protected. Given any individual $i$, a social choice $\mathcal{C}$ is said to respect $i$’s individual rights if, for all preference profiles $\rho \in \mathcal{R}$,
\begin{equation}
    [(\sigma_1,\sigma_2)\in D_i \, \, \& \, \, \sigma_1 \succ_i \sigma_2] \Rightarrow \sigma_2 \notin \mathcal{C}(\rho)
\end{equation}
With this definition of protecting individual preferences or rights, Sen proposed two axioms:

\begin{itemize}
    \item {\bf Pareto:} $\mathcal{C}$ satisfies Pareto if, for all preference profiles $\rho \in \mathcal{R}$, if $\sigma_1 \succ_i \sigma_2$ for all $i\in \mathcal{N}$, then $\sigma_2\notin \mathcal{C}(\rho)$.
    \item {\bf Minimal Liberalism:} A choice correspondence $\mathcal{C}$ respects minimal liberalism if there are at least two individuals, $i, j \in \mathcal{N}$, for whom $\mathcal{D}_i \neq \emptyset$ and $\mathcal{D}_j \neq \emptyset$, such that $\mathcal{C}$ respects both $i$’s and $j$’s individual rights.
\end{itemize}

\begin{theorem}[Sen's Theorem]
There exists no choice correspondence $\mathcal{C}$ satisfying universal domain, minimal liberalism, and Pareto.
\end{theorem}

\noindent Sen's theorem implies that any choice $\mathcal{C}$ satisfying universal domain and Pareto can respect the individual rights of at most one individual. The theorem demonstrates that protecting the preferences of multiple individuals is at odds with the most basic notions of utilitarian ethics. 

\medskip

\noindent To understand the application of Sen's theorem to AI alignment, consider a modified version of the scenario presented in his original paper~\cite{sen1970impossibility}.\footnote{In the original paper, Sen described the case of two individuals in a society - Lewd and Prude - who vote on whether they should or shouldn't be allowed to read Lady Chatterly's Lover.} Consider two reinforcers, A and B, and three outputs from an AI model (in the pre-RLHF stage). The three outputs correspond to normative statements about a political party X:

\begin{itemize}
    \item {\bf Output-1:} Political party $X$ is fascist
    \item {\bf Output-2:} Political party $X$ is not fascist
    \item {\bf Output-3:} Political party $X$ has a complicated agenda but it does not neatly fall into the above two categories. 
\end{itemize}

\noindent  Reinforcer $A$, who is very anti-$X$, prefers output 1, but given the choice between revealing their political bias and staying neutral, they would prefer to stay neutral over revealing their political affiliation. In decreasing order of preference, their ranking is 3, 1, 2. Reinforcer-$B$, however, does not mind revealing their political beliefs and would rather assert Output 2 over staying neutral. Their ranking is: 2, 3, 1.

\medskip

\noindent If the choice is between Outputs 1 and 3, a liberal - someone who cares about individual rights above all else - might argue that Reinforcer-$B$'s preference should count; since Reinforcer-A would be OK not revealing their preference, and should not be forced to. Thus the final output after RLHF would lead to Output-3. 

\medskip

\noindent Similarly, in the choice between Output-2 (``X is not fascist") and Output-3 (``its complicated"), liberal values require that Reinforcer-B's preference should be decisive, and since they clearly feel strongly about opposing the stance that X is fascist, they should be permitted to do this. In this case, Output-2 should be judged as ``socially acceptable" better than Output-3. 

\medskip

\noindent Thus, respecting individual preferences or liberal values would lead to preferring Output-3 over Output-1 and Output-2 over Output-3. Choosing the outputs from only one Reinforcer will break the Pareto axiom. We are thus left with an inconsistency.

\medskip

\noindent Note that unlike Arrow's theorem, Sen's theorem does not require  the transitivity axiom, or the independence of irrelevant alternatives. It only requires the existence of a best alternative (the Pareto axiom). Sen's result is thus stronger because it relies on fewer axiomatic constraints.

\section{Implications for AI Governance and Policy}
\label{sec:implications}
Can we build aligned AI agents using democratic norms?
The results in the previous section show the answer is not straightforward. Arrow's impossibility theorem implies that there is no unique voting rule that can allow us to train AI agents through RLHF while respecting democratic norms i.e., treating each reinforcer equally. Note that this is distinct from the obvious result that any democratic process will always result in a minority that will be unhappy about the outcome; however, Arrow's theorem implies that no unique voting rule exists even when deciding what the majority preference is. 

\medskip

\noindent Sen's theorem has even more serious implications for AI alignment using RLHF. It implies that a democratic method of aligning AI using RLHF cannot let more than one reinforcer encode their (privately held) ethical preferences via RLHF, irrespective of the ethical preferences of other reinforcers. As Sen argued in his original paper, ``liberal values conflict with the Pareto principle." We now discuss some policy implications of the Arrow-Sen results for RLHF. 

\subsection{Non-Uniqueness of AI Alignment and Model Transparency}

\noindent Consider multiple AI model developers/companies who want to build aligned AI agents. The choice of voting rule for each developer is private, i.e., there is no sharing of voting protocol between the developers. Even when the preferences of the reinforcers hired by the different developers are consistent, the resulting AI models might not be consistently aligned since voting rules across developers might differ. This will likely pose challenges for an alignment audit for AI models unless model builders disclose voting rules for the reinforcers. 

\medskip

\noindent One solution to this problem is to include the voting rule in a model card~\cite{Mitchell_2019}. This will allow consistent comparisons between different RLHF models. However, to avoid confusion for users the simplest solution is for all AI model builders to agree on a specific voting rule. It is not obvious how this voting rule can be agreed upon and poses a new governance challenge for AI alignment. Currently, model developers like OpenAI do not report voting rules in their system cards~\cite{openai2023gpt4}.

\subsection{Universal vs. Narrow AI Alignment}

\noindent Sen's theorem has important implications for aligning AI systems that respect personal preferences of individual users or a small subset of users with shared values, especially those who might not be represented amongst reinforcers. Consider, for example, an AI conversational agent trained via RLHF to be used in educational settings. Each user will likely have individual (private) preferences that other users or model developers should not have any say on. For example, a user of an AI conversational agent in a classroom might prefer not to encounter language deemed racist by them. Sen's theorem implies that it is impossible to build an RLHF model via democratic methods, such that every individual's private preferences are respected. While {\it prima facie}, this conclusion might seem obvious, it implies that we cannot build artificial general intelligence aligned with all users' intentions. Indeed {\it any} AI agent built using RLHF will be misaligned with every user in some dimension. 

\medskip

\noindent This result has important implications for developing universally aligned AI agents. At the outset, building universally aligned AI agents using RLHF is impossible. We can build AI agents that are narrowly aligned with certain preferences of a group of users but these preferences need to be explicitly communicated during the reinforcement process. For example, we can build an AI conversational agent that avoids racist speech (suitably defined). But we cannot expect the agent to be aligned along other dimensions, such as gender, nationality, political background, etc., that were not explicitly considered during the training process. This implies we can build aligned AI agents for narrow preferences/domains. In practice, this means we will have a family of aligned models for specific tasks/groups but not generally aligned models.

\section{Conclusion}
\label{sec:conclusion}
The development of a universally accepted ethical framework for building and regulating AI is still a work in progress. However, there has been  convergence in key principles that communities around the world want in AI systems: transparency, fairness, non-maleficence, responsibility and privacy~\cite{jobin2019global}. There is however no consensus between entities developing and regulating AI around how to embed these principles into AI models. 

\medskip

\noindent RLHF has shown remarkable success in improving model outputs of LLMs. However, whether it is possible to align models in a scalable manner, while respecting democratic norms, remains unclear. This paper argues that even in the presence of diverse reinforcers who are representative of a  population of users, alignment might not be unique. Further, we argue that there will always be private preferences of users that an RLHF AI model built via democratic norms will violate. 

\medskip

\noindent Our results have important implications for the governance and development of aligned AI systems. While the use of RLHF for real-world applications has exploded in recent months, regulations and policies for deploying RLHF models at scale are still nascent. Our results highlight the collective action challenges that model developers and policymakers will have to grapple with while aligning AI agents in the near future.
Finally, our results indicate that the future of aligned AI development might be better served by incentivizing smaller model developers working on aligning their models to a narrow set of users (``narrow AI alignment") as opposed to trying to build universally aligned AI (``aligned AGI").

\printbibliography[]

\end{document}